\documentclass{article}

\usepackage[final]{nips_2016}

\usepackage[utf8]{inputenc} % allow utf-8 input
\usepackage[T1]{fontenc}    % use 8-bit T1 fonts
\usepackage{nicefrac}       % compact symbols for 1/2, etc.
\usepackage{microtype}      % microtypography
\usepackage{amsmath,amssymb,amsthm,amsfonts}
\usepackage{caption,subcaption}
\usepackage{times}
\usepackage{xr}
\usepackage{hyperref}
\usepackage{url}
\usepackage{amssymb}
\usepackage{amsbsy}
\usepackage{amsthm}
\usepackage{amsmath}
\usepackage{paralist}
\usepackage{bm}
\usepackage{booktabs}
\usepackage{rotating}
\usepackage{cases}
\usepackage{mathtools}
\usepackage{graphicx}
\usepackage{algorithm}
\usepackage{algpseudocode}
\usepackage{setspace}
\usepackage{textcomp}
\usepackage{graphicx}
\usepackage{multirow}
\usepackage{color}

\title{Diet2Vec: Multi-scale analysis of massive dietary data}

\author{
  Wesley Tansey \\
  Department of Computer Science \\
  University of Texas at Austin \\
  Austin, TX 78712 \\
  \texttt{tansey@cs.utexas.edu} \\
  \And
  Edward W. Lowe, Jr. \\
  FitNow, Inc. \\
  Boston, Massachusetts 02210 \\
  \texttt{will@loseit.com} \\
  \AND
  James G. Scott \\
  Department of Information, Risk, and Operations Management \\ Department of Statistics and Data Sciences \\
  University of Texas at Austin \\
  Austin, TX 78712 \\
  \texttt{james.scott@mccombs.utexas.edu} \\
}

\begin{document}

\maketitle

\begin{abstract}
Smart phone apps that enable users to easily track their diets have become widespread in the last decade. This has created an opportunity to discover new insights into obesity and weight loss by analyzing the eating habits of the users of such apps. In this paper, we present diet2vec: an approach to modeling latent structure in a massive database of electronic diet journals. Through an iterative contract-and-expand process, our model learns real-valued embeddings of users' diets, as well as embeddings for individual foods and meals. We demonstrate the effectiveness of our approach on a real dataset of 55K users of the popular diet-tracking app LoseIt\footnote{http://www.loseit.com/}. To the best of our knowledge, this is the largest fine-grained diet tracking study in the history of nutrition and obesity research. Our results suggest that diet2vec finds interpretable results at all levels, discovering intuitive representations of foods, meals, and diets.
\end{abstract}

% !TEX root = diet2vec.tex
\section{Introduction}
\label{sec:introduction}
Historically, nutritional research has been limited to one of two scenarios. To gather large-scale datasets, researchers were limited to high-level epidemiological surveys and questionnaires that only gathered coarse-grained information. For example, researchers may ask questions such as ``How often have you gone on a diet during the last year?''  \citep{neumark:etal:2006}, which fails to capture in-depth information on specific foods and time frames. Gathering fine-grained data, such as daily food journals, was only possible in small sample sizes due to the prohibitively high cost of recruiting and tracking subjects. Even the most ambitious of projects to-date only track subjects infrequently via random check-ins \citep{gardner:etal:2007}. 

But the advent of smart phones and applications for diet tracking has created a third category of nutritional data.  The millions of users who track their diets on apps are doing so at a fine-grained level, providing both the name and the nutritional information for everything they have eaten.  They also provide information about meals: i.e.,~which foods they have eaten together.  This gives us a powerful source of information about which foods tend to be eaten together.  Essentially, this data combines the best of both worlds: it exceeds the scale of even the largest epidemiological surveys, but also provides the fine-grained detail of a food journal.

However, the massive amounts of data created by these users is often noisy and error-prone, which distinguishes it from typical data collected in small-scale food-journaling studies.  Most foods in diet tracking app databases are crowd-sourced, and therefore often contain missing or incorrect information in the form of misspelled food names and incorrect nutritional-content entries. The resulting datasets generated by these apps hold a wealth of knowledge, but require more sophisticated analysis methods to handle the complexity and noisiness of the underlying data.

\begin{figure}[t]
\centering
\begin{subfigure}[t]{.45\textwidth}
\frame{\includegraphics[trim={2.9cm 1.5cm 2.9cm 1.45cm},height=2.5in]{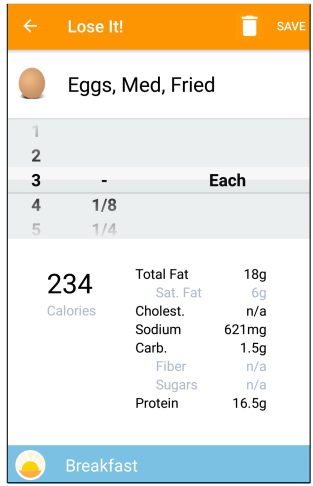}}
\end{subfigure}
\begin{subfigure}[t]{.45\textwidth}
\includegraphics[width=\textwidth]{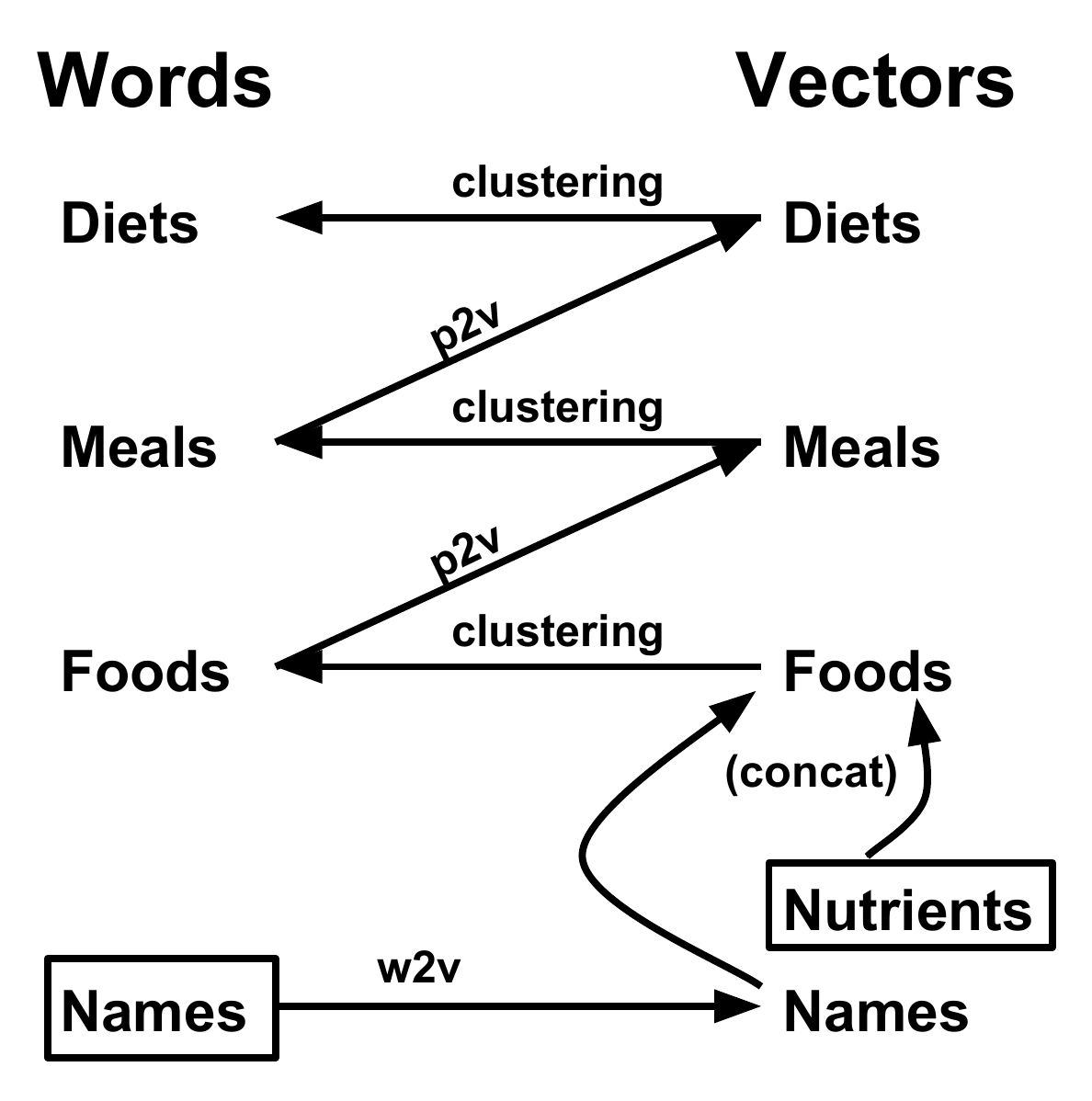}
\end{subfigure}
\caption{\label{fig:diet2vec_overview} An overview of the diet2vec system. We build both vector embeddings (via word2vec and paragraph2vec) and interpretable clusters in a bottom-up approach. Boxed entries correspond to user-specified data about each food.}\vspace{-0.2in}
\end{figure}

In this paper, we present diet2vec, a scalable, robust approach for modeling nutritional diaries from smart phone apps. We start from the bottom up: that is, with a collection of food entries, each containing a name and a vector of real-valued nutrients. For example, Figure \ref{fig:diet2vec_overview} (left) shows an example food log entry of three fried eggs, with the name specified at the top and the nutrients at the bottom. Figure \ref{fig:diet2vec_overview} (right) shows an overview of the diet2vec approach. At each level diet2vec learns a vector embedding of the current available word representation in the previous level. Those vectors are then clustered to form the ``words'' for the next level above, where the process is repeated. The result is a multi-scale view of nutrition at the food, meal, and diet levels, with both interpretable clusters and real-valued embeddings that may be useful for downstream modeling applications. We used diet2vec to model 55K active users on the LoseIt smart phone app, making this the largest fine-grained diet analysis ever conducted.

% !TEX root = diet2vec.tex
\section{Modeling Foods}
\label{sec:modeling_foods}
Every user of the diet app is able to log the foods they eat every day. Each food log entry specifies which food the user ate, how much of it was consumed, and a meal tag (breakfast, lunch, dinner, or snacks). Each food log entry is an instance of a unique food entry from a database of crowd sourced foods, each containing a name of the food in plain text and a vector of macro-nutrients (i.e., fat, carbs, and protein) and micro-nutrients (saturated fat, cholesterol, sodium, fiber, and sugar) for that food. We note that the crowd sourced nature of the foods make the data particularly difficult to deal with: there are many overlapping entries, often entries are missing some nutrient information, and names are frequently misspelled. Foods are also sparsely reused: our 55K users generated roughly 88M food log entries composed of 4.5M unique foods. Thus, we need to perform dimensionality reduction in a way that leverages the structure of each food.

To model foods, we first run word2vec \citep{mikolov:etal:2013} on the names of the foods, treating each food name as a separate document. We then average the vectors for each food name's component vectors to arrive at a food name vector. The nutrient component of each food is converted to a per-calorie measure to maintain portion invariance. The food name vectors are then concatenated with the nutrient vectors, standardized by median deviations, and winsorized at +/- 2.5 deviations to produce the final food vectors. This last step is necessary to handle the large per-calorie outliers of certain foods.  A good example here is vitamins, which may have very low calories but high nutrient content.

\begin{figure}[t]
\centering
\begin{subfigure}[t]{.32\textwidth}
\caption{cheddar shredded; cheese mild}
\includegraphics[width=\textwidth]{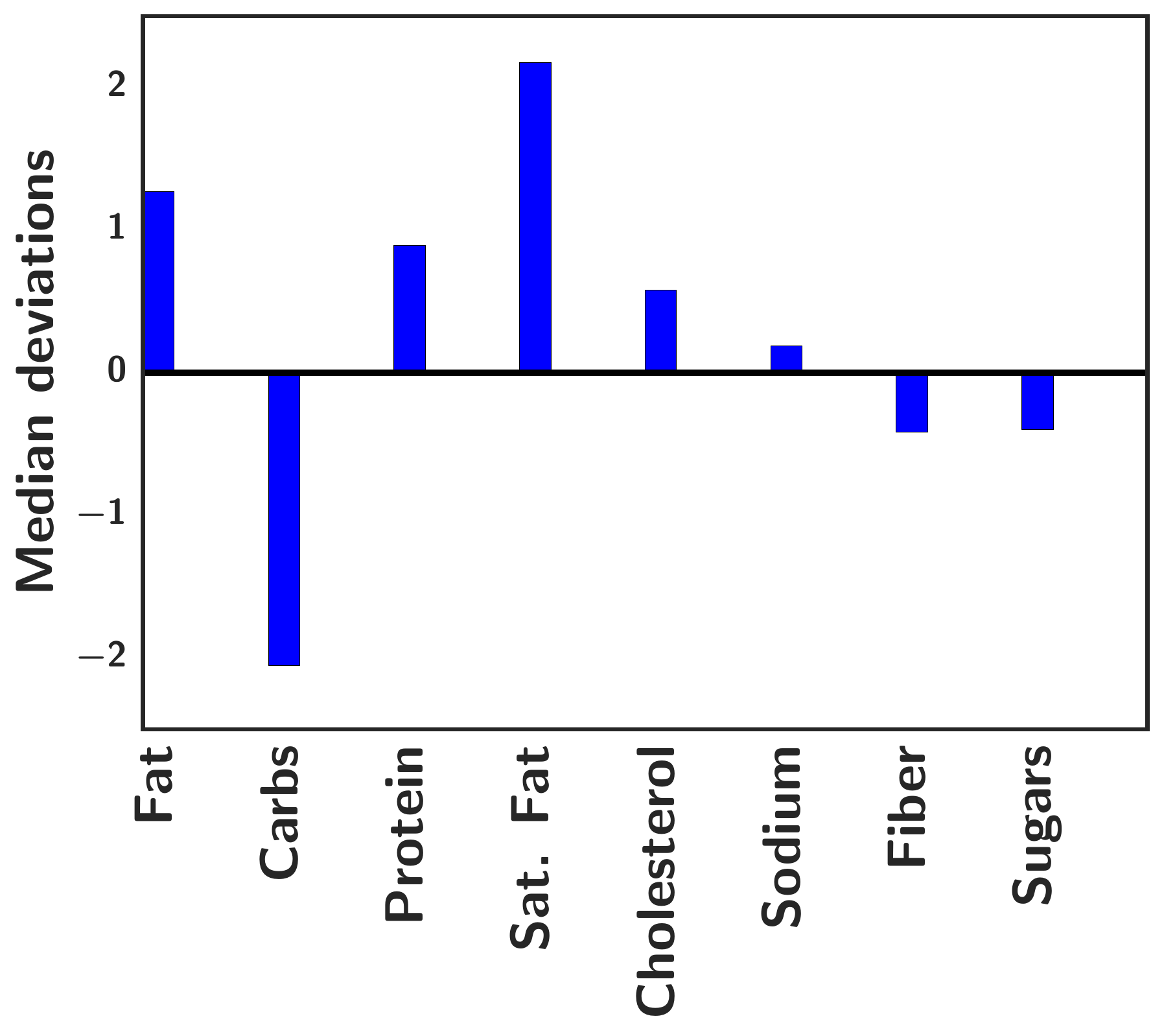}
\end{subfigure}
\begin{subfigure}[t]{.32\textwidth}
\caption{ground chuck; beef burgers}
\includegraphics[width=\textwidth]{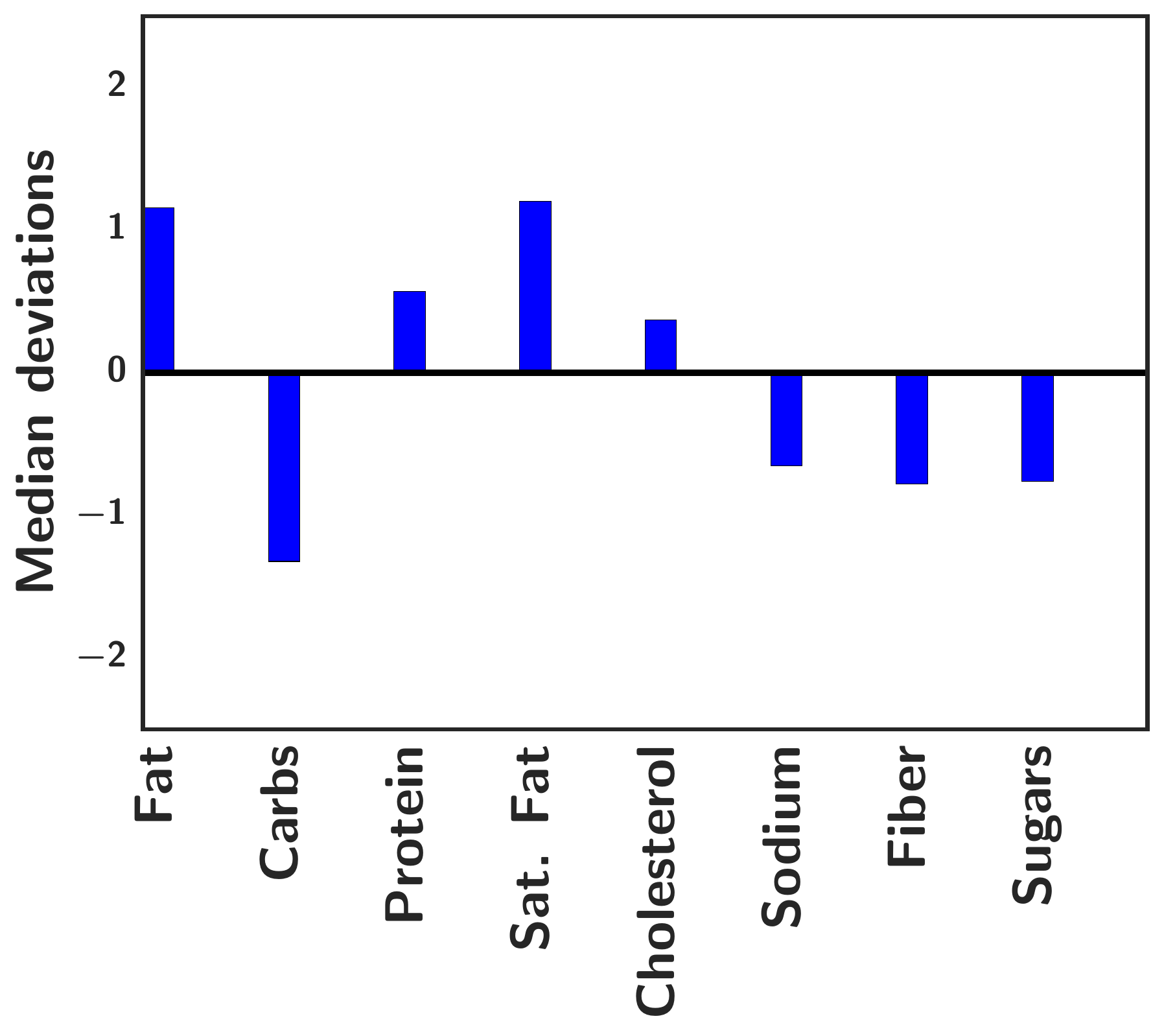}
\end{subfigure}
\begin{subfigure}[t]{.32\textwidth}
\caption{hot cocoa; chocolate mix}
\includegraphics[width=\textwidth]{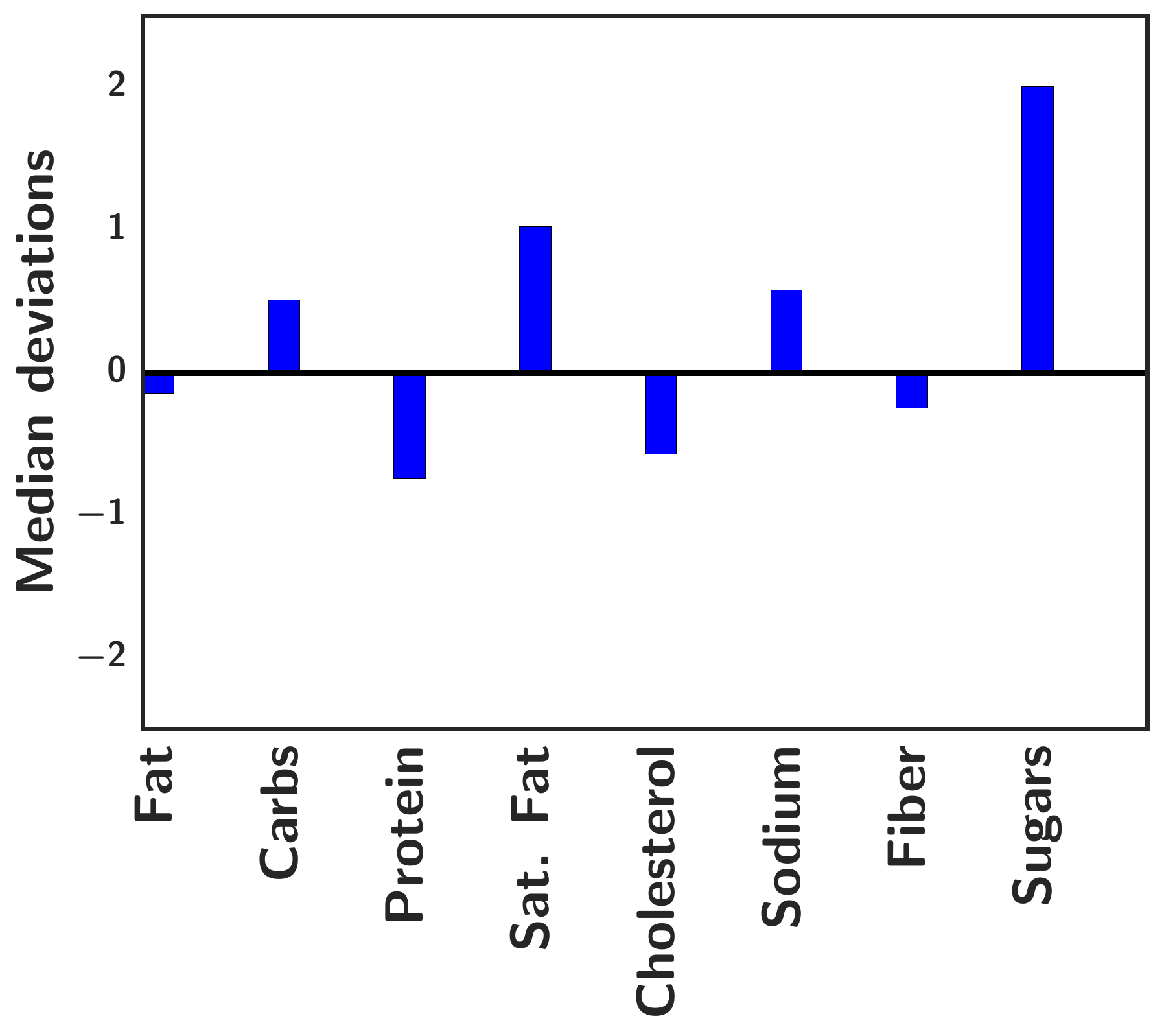}
\end{subfigure}
\caption{\label{fig:food_cluster_examples} Examples of three food clusters discovered by our model. For each food, we show the nutrient portion of its centroid and its name as automatically-derived via TF-IDF on the name terms. Each nutrient is measured in terms of median deviations on a per-calorie basis, with 0 being exactly the median. The food clusters correspond well to their names: the cheese and ground beef clusters have high fat and protein, while the hot cocoa cluster is mostly sugar.}
\end{figure}

Once vector representations of foods have been created, we run weighted k-means to cluster the foods into 5000 ``food words'', placing $20\%$ of the weight on the name and $80\%$ of the weight on the nutrients. As noted earlier, many foods are very similar but may have many duplicate or similar entries. The resulting clusters are then automatically named according to their top n-grams ($n=\{1,2\}$) ranked by TF-IDF. Three example clusters are visualized in Figure \ref{fig:food_cluster_examples}. Each cluster centroid's nutrient values correspond intuitively to their cluster names, producing highly interpretable clusters.

% !TEX root = diet2vec.tex
\section{Modeling Meals}
\label{sec:modeling_meals}
Food log entries in most diet tracking apps, including LoseIt, are grouped into meals. Conceptually, a meal can be seen as a set of foods. Given our food words from section \ref{sec:modeling_foods}, we next generate meal vectors via the DBOW model of paragraph2vec \citep{le:mikolov:2014}. As with foods, we then cluster the meal vectors to get ``meal words.'' For our dataset, users logged approximately 24M meals which we clustered into 1000 words.

Table \ref{tab:meals} shows some examples of meal words, with their associated top food words by TF-IDF. Again, the results are easily interpreted, with some clearly pointing to a type of food (e.g. Mexican food in Meal 1), a generic meal type (an American-style breakfast of eggs and bacon in Meal 2), a common meal combo (Subway's sub, chips, and cookie combo deal in Meal 3), common ingredients in a generic meal (e.g., sandwich in Meal 4, salad in Meal 6), and popular snack combinations (kale or tortilla chips with hummus in Meal 5). Importantly, unlike the food vectors, the meal vectors are based on co-occurrence. This enables the meal words to contain foods that are very different nutritionally but pair well in a meal.

\begin{table}[thb]
\centering
\hspace{-0.3in}\begin{tabular}{|c|l|l|l|}
\hline
& \multicolumn{1}{c|}{Meal 1} & \multicolumn{1}{c|}{Meal 2} & \multicolumn{1}{c|}{Meal 3} \\
\hline
\parbox[t]{2mm}{\multirow{5}{*}{\rotatebox[origin=c]{90}{Top Foods}}} & relleno, flautas & italian dry, bacon cooked & chip soft, cookie chocolate \\
& grain wild, ready rice & hash browns, browns & raisin cookies, cookie oatmeal \\
& great northern, beans black & egg scrambled, eggs scrambled & cookies sandwich, oreo cookie \\
& carne asada, asada & milk vitamin, vitamin d & subway turkey, subway club \\
& chimichanga, taquitos & skim vitamin, milk skim & potato crisps, chips baked \\
\hline
\noalign{\smallskip}\noalign{\smallskip}
\hline
& \multicolumn{1}{c|}{Meal 4} & \multicolumn{1}{c|}{Meal 5} & \multicolumn{1}{c|}{Meal 6} \\
\hline
\parbox[t]{2mm}{\multirow{5}{*}{\rotatebox[origin=c]{90}{Top Foods}}} & ham honey, lunchmeat turkey & kale chips, seaweed & sweet peppers, super sweet\\
& cheese muenster, cheese blue & pepper hummus, hommus & garbanzo beans, beans kidney \\
& bread honey, bread wheat & hummus classic, sabra & sugar snap, beans green \\
& spicy brown, mustard yellow & hommus, humus & grilled o, skinless cooked\\
& miracle whip, miracle & super sweet, corn sweet & spinach baby, baby fresh\\
\hline
\noalign{\smallskip}\noalign{\smallskip}
\end{tabular}
\caption{\label{tab:meals}Example meal clusters found by our algorithm. For each meal, we list the top five food clusters by TF-IDF. The clusters are highly interpretable, with an intuitive semantic meaning to each group of foods.}
\end{table}

% !TEX root = diet2vec.tex
\section{Modeling diets}
\label{sec:modeling_diets}
\begin{figure}[t]
\centering
\begin{subfigure}[t]{.32\textwidth}
\includegraphics[width=\textwidth]{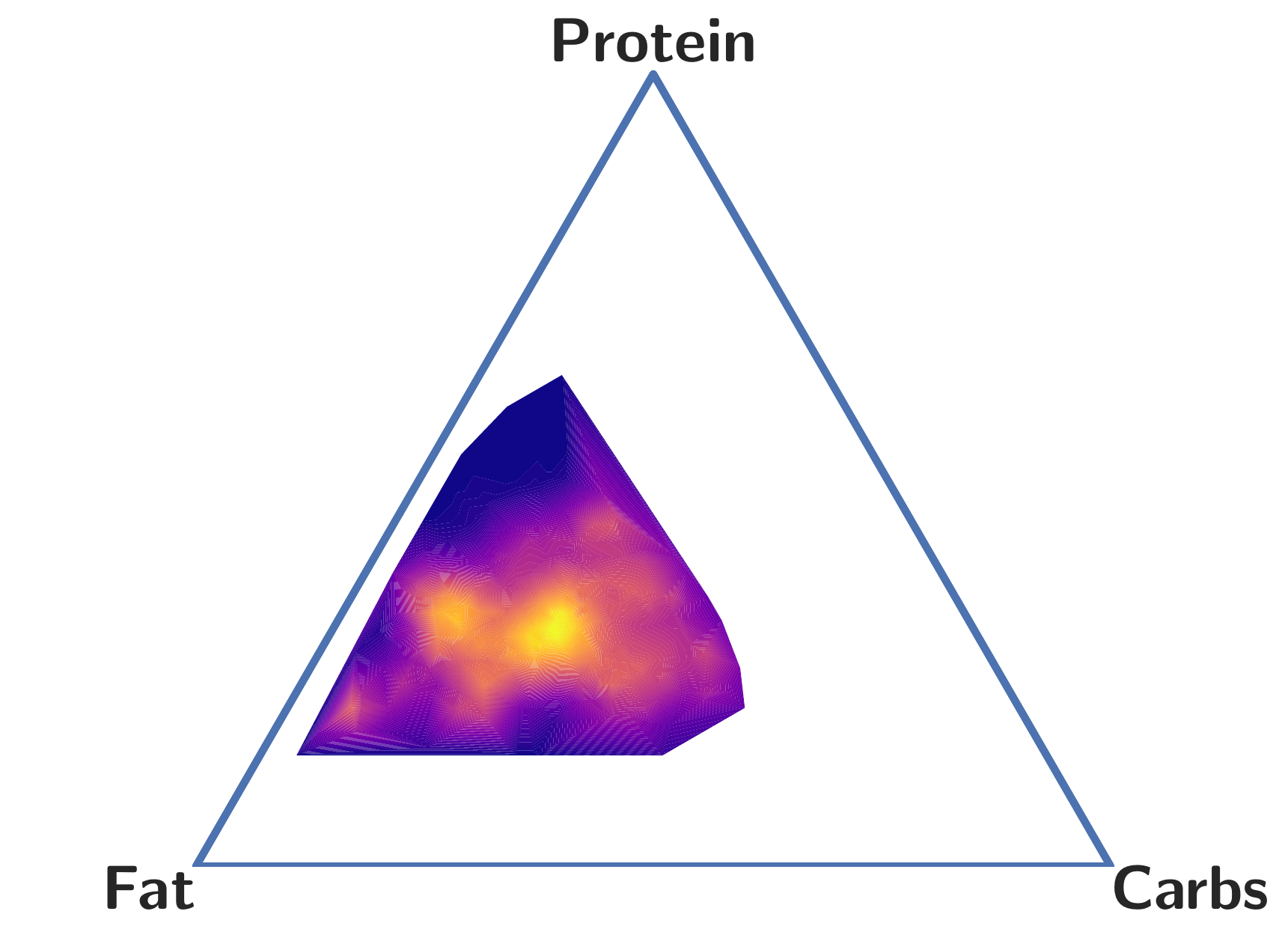}
\end{subfigure}
\begin{subfigure}[t]{.32\textwidth}
\includegraphics[width=\textwidth]{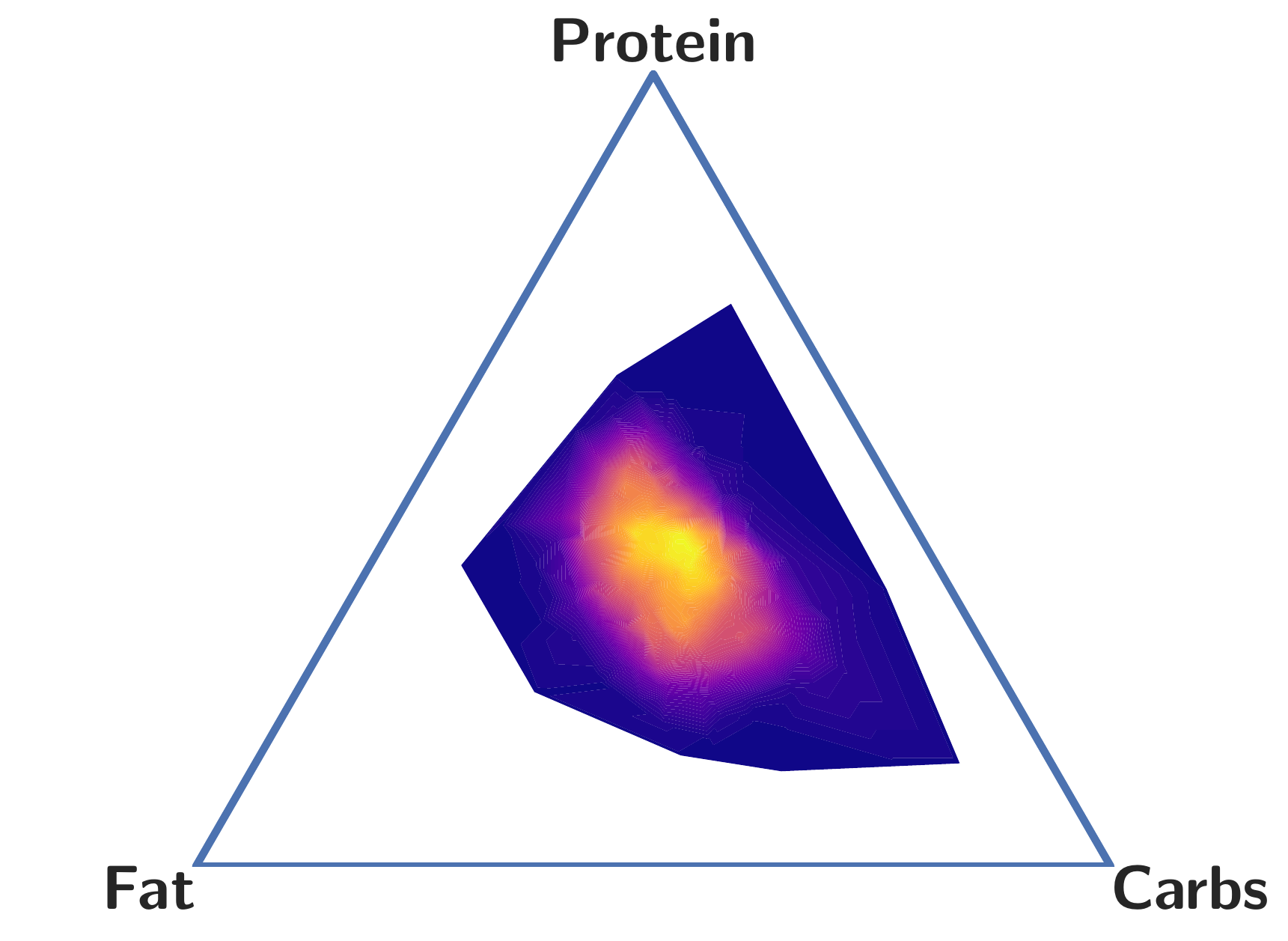}
\end{subfigure}
\begin{subfigure}[t]{.32\textwidth}
\includegraphics[width=\textwidth]{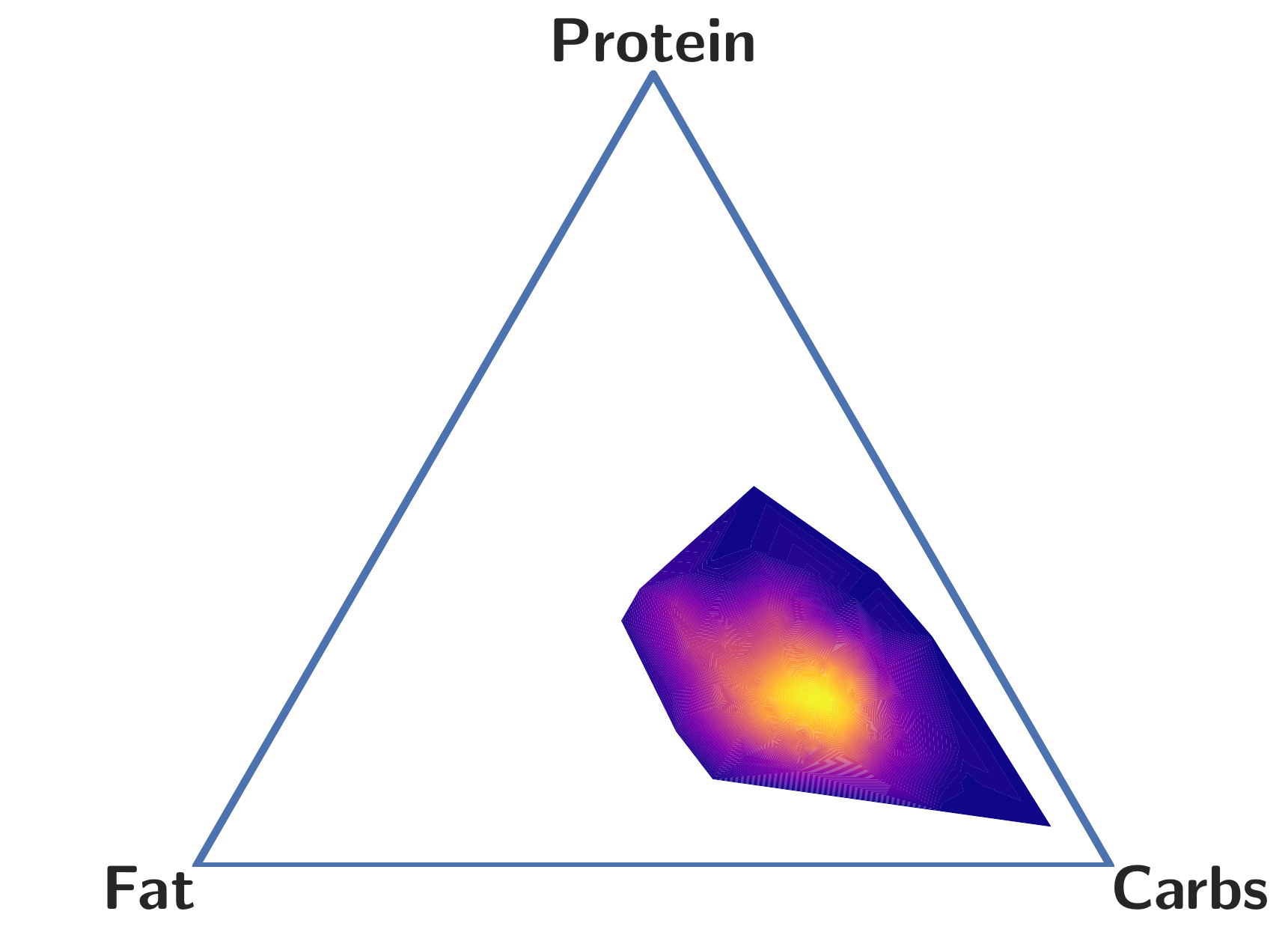}
\end{subfigure}
\caption{\label{fig:diet_clusters} Three example diet clusters plotted as a function of their constituent users' average daily macro ratios. Despite the vectors being generated based on meal co-occurrence, our resulting diet vectors still have intuitive interpretations in terms of macro ratios. The left cluster can be seen as a ``low-carb'' diet group, the right as ``high-carb, low-fat'', and the middle as more balanced.}
\end{figure}

We represent each user's diet as a bag of meal words and again generate diet vectors. The vectors are then clustered into 100 diet words, which can conceptually be considered groups of people who eat similarly. Such groupings are useful for several downstream applications, including:
{\setstretch{0.5}
\begin{enumerate}
    \item Recommending meals and food substitutions.
    \item Predicting user weight loss and churn.
    \item Tailoring support messages to encourage users to continue their diet tracking.
\end{enumerate}}
Perhaps surprisingly, our diet clusters have meaningful interpretations as distributions over macro ratios. Even though we trained on meal words, which are specifically designed not to include any nutritional information, the meals themselves often correspond to certain nutritional profiles. Figure \ref{fig:diet_clusters} shows three examples of our diet clusters. Each figure shows the daily macro ratios for the users that were assigned to that cluster. The left group is clearly a low-carb cluster, while the right group is clearly very high-carb; the middle cluster represents a more balanced diet of fat, protein, and carbs.

% !TEX root = diet2vec.tex
\section{Conclusion}
\label{sec:conclusion}
We presented diet2vec, a method for analyzing massive amounts of nutritional data generated by users of diet tracking apps. Our method produces a multi-scale view of the data by analyzing logs at the food, meal, and diet levels. Each higher level leverages the output of the lower level to recursively build a hierarchy of highly-interpretable clusters. We also showed that our final diet clusters have semantic meaning in terms of their component users' average daily macro ratios. Finally, we wish to stress that all results for the food, meal, and diet clusters are not ``cherry-picked''---they are broadly representative of typical clusters, with uninterpretable groups being the exceptions in each case. The diet2vec model is the first approach to large-scale modeling of fine-grained diet information of which we are aware. We believe the interpretable output of the model will empower obesity and nutrition researchers to answer many new questions, as well as lending new insights into long-standing questions about diet efficacy and eating behaviors.

\begin{small}
\bibliographystyle{abbrvnat}
\bibliography{diet2vec}
\end{small}

\end{document}